\documentclass{article}
\usepackage[preprint]{neurips_2026}
\usepackage[utf8]{inputenc}
\usepackage[T1]{fontenc}
\usepackage{hyperref}
\usepackage{url}
\usepackage{booktabs}
\usepackage{longtable}
\usepackage{amsmath}
\usepackage{amsfonts}
\usepackage{amssymb}
\usepackage{nicefrac}
\usepackage{microtype}
\usepackage{xcolor}
\usepackage{graphicx}
\usepackage{tikz}
\usetikzlibrary{arrows.meta, positioning, shapes.geometric, calc, fit, backgrounds}

\definecolor{real-fill}{HTML}{D4EDDA}
\definecolor{real-draw}{HTML}{28A745}
\definecolor{farl-fill}{HTML}{CCE5FF}
\definecolor{farl-draw}{HTML}{004085}
\definecolor{dino-fill}{HTML}{E8F4FD}
\definecolor{dino-draw}{HTML}{0066CC}
\definecolor{dca-fill}{HTML}{FFF3CD}
\definecolor{dca-draw}{HTML}{856404}
\definecolor{rfg-fill}{HTML}{E2D9F3}
\definecolor{rfg-draw}{HTML}{5A189A}
\definecolor{arrow-gray}{HTML}{555555}

\title{Dimensional Coactivation for Representational Consistency\\
in Frozen Vision Foundation Models}

\author{
  Izaldein Al-Zyoud \quad Abdulmotaleb El Saddik\\
  MCRLab, School of Electrical Engineering and Computer Science\\
  University of Ottawa, Ottawa, ON, Canada\\
  \texttt{izzy.alzyoud@uottawa.ca}
}

\begin{document}
\maketitle

\begin{abstract}
Frozen vision foundation models do not merely extract features; they organize images through a learned coordinate system. We ask whether that coordinate system remains internally coherent within a single input. This leads to \textbf{Representational Consistency}: the study of whether a frozen foundation model represents one sample coherently across its semantic subregions.

We introduce \textbf{Dimensional Coactivation} (DCA), a per-dimension instrument for measuring this coherence. DCA compares semantic regions by asking whether the same feature dimensions coactivate across them. Unlike classical similarity measures, it deliberately avoids centering, $L_2$ normalization, and full Gram coupling. These operations are useful when comparing different models or distributions, but they are mismatched to the intra-sample setting, where the coordinate system is fixed and raw magnitude carries signal.

Deepfake detection provides a natural validation task. Synthetic faces may reproduce plausible eyes, noses, and mouths while breaking the representational structure that links those regions in real faces. Using frozen DINOv3 features, DCA exposes this break: an eyes-mouth-nose fingerprint achieves 0.9106 AUC on CelebDF-v2 and 0.9289 on DFD under FF++ c23 cross-dataset transfer. The design is also sharply validated by ablation: reintroducing centering collapses CelebDF-v2 AUC to 0.459, $L_2$ normalization reduces it to 0.862, and cross-dimension coupling reduces it to 0.478.

Finally, replacing DINOv3 with FaRL collapses CelebDF-v2 AUC to 0.582. DCA therefore depends on a stable per-dimension coordinate system, not on region extraction alone. These results position DCA as an instrument for measuring intra-sample representational coherence in frozen foundation models, with deepfake detection as the first validation task.
\end{abstract}

\section{Introduction}
\label{sec:intro}

A frozen vision foundation model does more than extract features from an image. It imposes a coordinate system: a set of feature dimensions through which visual structure is organized. If this coordinate system is stable, then different semantic regions of the same input should not be represented as independent fragments. The eyes, nose, and mouth of a real face, for example, should remain coherent inside the model because they belong to the same physical identity.

This paper studies that property. We call it \emph{Representational Consistency} (RC): the study of whether a frozen foundation model represents a single input as internally coherent across its semantic subregions. RC introduces an axis of representation measurement that is distinct from standard similarity frameworks. CKA and RSA compare representations across models or layers; OOD methods compare a test sample against a training distribution; multi-view consistency compares augmented views of the same content during training. RC instead compares region $R_1$ and region $R_2$ inside the same input, under the same frozen model, at inference time. It is therefore an intra-sample, intra-model, inter-region measurement problem.

Deepfake detection provides a natural validation task for this idea. A synthetic face may reproduce locally plausible eyes, noses, and mouths while failing to preserve the representational structure that links those regions in real faces. In that case, the image may appear realistic region by region, but still be internally inconsistent inside the feature space of a model trained on physical-world imagery. The goal is not to build a detector by fine-tuning a new network, but to ask whether a frozen foundation model already contains a measurable coherence residual that exposes this failure.

We instantiate RC with \emph{Dimensional Coactivation} (DCA), a per-dimension measurement instrument for cross-region coherence. Given two semantic regions, DCA asks whether the same feature dimensions coactivate across both regions. For the eyes-mouth-nose region triple, the three pairwise DCA vectors form a 3{,}072-dimensional fingerprint. This fingerprint is not a pooled regional descriptor and not a scalar similarity score; it preserves the feature-dimension axis where the frozen model's coordinate system is readable.

The design of DCA follows from the measurement regime. Classical similarity measures use centering, $L_2$ normalization, and full Gram coupling because they compare representations across different models, layers, or distributions, where means, scales, and coordinate systems may not align. RC compares regions inside one frozen model, where the coordinate system is fixed. In this setting, mean activation is not necessarily a nuisance bias; it may encode identity coherence. Magnitude is not arbitrary scale; it may carry identity load. Off-diagonal cross-dimension interactions are not guaranteed to be stable and can obscure the diagonal structure. We refer to removing these mismatched assumptions as \emph{constraint liberation}.

Empirically, training a linear probe on FF++ c23 and evaluating zero-shot on CelebDF-v2 and DFD, the eyes-mouth-nose DCA fingerprint achieves 0.9106 AUC on CelebDF-v2 and 0.9289 on DFD. Reintroducing centering, $L_2$ normalization, or full Gram coupling each damages the signal. Replacing DINOv3 with FaRL collapses CelebDF-v2 AUC to 0.582, which diagnoses the absence of the per-dimension coordinate system that RC measurement requires.

Our contributions are fourfold. First, we introduce Representational Consistency as an intra-sample, intra-model, inter-region measurement problem for frozen foundation models. Second, we propose DCA as a per-dimension operator that measures whether the same feature dimensions coactivate across semantic regions of one input. Third, we show that constraint liberation is necessary for this measurement regime, with independent ablations for centering, normalization, and cross-dimension coupling. Fourth, we validate the measurement through cross-dataset deepfake detection and show that its success depends on the coordinate-system stability of the frozen backbone.

\section{From Similarity to Representational Consistency}
\label{sec:related}

Representational Consistency is closest to the literature on representation similarity, but it changes the object being measured. Classical tools such as CKA~\citep{kornblith2019cka}, RSA~\citep{kriegeskorte2008rsa}, SVCCA, PWCCA, and related alignment methods ask whether two representational spaces are similar across models, layers, random seeds, or training conditions. Their output is usually a scalar summary: a report of how close two representations are after removing differences in scale, offset, or basis. These invariances are appropriate when the comparison is across coordinate systems. They become less appropriate when the comparison is inside one frozen model, where the coordinate system is shared by construction.

RC occupies this intra-model regime. It does not ask whether two models represent the same dataset similarly, or whether two layers of a network are aligned. It asks whether two semantic regions of the same input remain coherent inside one frozen model. This distinction changes what should be preserved. For inter-model comparison, centering and normalization make the comparison fair. For intra-sample comparison, those same operations can delete the signal: the mean activation and magnitude of a feature dimension may be exactly what encodes cross-region coherence.

This difference also separates RC from multi-view consistency. In self-supervised learning, methods such as DINO~\citep{caron2021dino}, BYOL, and VICReg encourage different augmented views of the same image to produce compatible representations. Multi-view consistency is therefore a training-time objective over different views of the same content. RC is an inference-time measurement over different semantic content inside the same input. Eyes and mouth are not augmentations of one another; they are distinct regions that should nevertheless remain coordinated if they belong to the same physical identity. In this sense, RC can be understood as measuring a property that self-supervised training may induce, but does not explicitly target.

Second-order feature methods provide another nearby comparison point. Bilinear pooling, iSQRT-COV, and style-transfer Gram matrices~\citep{gatys2016style} use feature outer products to capture richer interactions than first-order pooled descriptors. Gram-based OOD methods~\citep{sastry2020detecting} similarly use second-order activation statistics to detect deviations from training behavior. These methods share a mathematical ingredient with DCA, namely products of feature activations, but they use it for a different purpose. They either aggregate the full cross-dimension structure into a scalar or learn a downstream classifier over coupled dimensions. DCA instead keeps only the same-dimension cross-region coactivation. The goal is not to model all pairwise feature interactions, but to preserve the diagonal axis where a stable frozen coordinate system can be read.

This is why DCA is better understood as a transducer rather than a comparator. A comparator consumes two representations and emits a similarity score. CKA asks: are these two representations globally similar? DCA consumes two regional activation matrices and emits a structured feature vector. It asks: which feature dimensions coactivate across semantic regions of the same input? The output is therefore not only reportable, but reusable as a sample-level fingerprint.

Table~\ref{tab:positioning} summarizes the distinction. The closest prior families each capture part of the required structure: some operate per-sample, some preserve magnitude, some use second-order products. DCA is the intersection needed for RC measurement: per-sample, per-dimension, magnitude-preserving, and regional.

\begin{table}[t]
\centering
\caption{Positioning DCA against closest prior second-order and similarity methods. Per-sample indicates one output per input. Per-dim indicates retention of the feature-dimension axis. Magnitude indicates that activation scale is preserved rather than normalized away. Regional indicates comparison across semantic subregions of one input.}
\label{tab:positioning}
\small
\begin{tabular}{lcccc}
\toprule
Method & Per-sample & Per-dim & Magnitude & Regional \\
\midrule
CKA / RSA & $\times$ & $\times$ & $\times$ & $\times$ \\
Gram-OOD & $\checkmark$ & $\times$ & scalar & $\times$ \\
Bilinear / iSQRT-COV & $\checkmark$ & $\times$ & trained & $\times$ \\
Gatys style-Gram & $\checkmark$ & $\times$ & $\checkmark$ & $\times$ \\
$\texttt{cosine\_dim}$ ($L_2$ ablation) & $\checkmark$ & $\checkmark$ & $\times$ & $\checkmark$ \\
\textbf{DCA} & $\checkmark$ & $\checkmark$ & $\checkmark$ & $\checkmark$ \\
\bottomrule
\end{tabular}
\end{table}

Deepfake detection methods have largely adapted the learned detector to the forensic task: stronger backbones~\citep{rossler2019ff,chollet2017xception}, self-blending~\citep{shiohara2022selfblending}, domain adaptation~\citep{haliassos2022leveraging}, attention mechanisms~\citep{zhao2021multiattention}, reconstruction~\citep{cao2022recce}, and region-specific modeling~\citep{haliassos2021lipforensics}. RC takes the opposite route: it keeps the backbone frozen and asks whether the pretrained representation already contains an intra-sample residual that distinguishes coherent from incoherent faces. Region-guided prior approaches use regions as places to pool, attend, reconstruct, or classify; DCA uses regions as anchors for a different measurement: whether the same feature dimensions coactivate across regions. The signal is relational, not regional.

Frozen self-supervised features have already been shown to contain dense geometric structure useful for correspondence across images~\citep{amir2022dino,tang2023emergent}. DCA moves the same idea to the intra-image axis: rather than asking whether a patch in one image corresponds to a patch in another, it asks whether semantic regions within one image remain coherent along the same feature dimensions. This makes DCA complementary to dense correspondence methods, not a replacement for them.

\section{Representational Consistency and Constraint Liberation}
\label{sec:rc_constraint_liberation}

Representational Consistency requires a different measurement regime from standard representation similarity. In CKA, RSA, and related methods, the comparison is usually across models, layers, random seeds, or training conditions. The two feature spaces may have different means, scales, and coordinate systems, so centering and normalization are useful invariances. They make the comparison fair by removing quantities that are treated as arbitrary.

RC compares something else: two semantic regions inside the same input and the same frozen model. In this setting, the coordinate system is already shared. The question is not whether two independently trained spaces can be aligned, but whether one frozen coordinate system remains coherent across regions of a single sample. This shift changes which quantities should be discarded and which should be preserved.

Given two semantic regions $r_1$ and $r_2$ with sampled patch-token matrices $F_{r_1}, F_{r_2} \in \mathbb{R}^{K \times D}$, DCA measures whether the same feature dimensions coactivate across both regions:
\begin{equation}
\label{eq:dca}
\mathrm{DCA}(F_{r_1}, F_{r_2})[d]
=
\frac{1}{K}
\sum_{k=1}^{K}
F_{r_1}[k,d]\,F_{r_2}[k,d],
\qquad d = 1,\ldots,D.
\end{equation}
Equivalently, DCA is the diagonal of the cross-region patch-token product, scaled by the number of sampled patches:
\[
\mathrm{DCA}(F_{r_1}, F_{r_2})
=
\frac{1}{K}\,\mathrm{diag}(F_{r_1}^{\top}F_{r_2})
\;\in\; \mathbb{R}^{D}.
\]

The operator is intentionally minimal. It keeps the feature-dimension axis, preserves activation magnitude, and discards cross-dimension interactions. These choices are not arbitrary simplifications. They follow from the fact that RC measures coherence within one frozen coordinate system rather than similarity across different coordinate systems.

A key design choice is to avoid collapsing cross-region coherence to a single scalar. Scalar summaries such as cosine similarity, CKA, or mean DCA can indicate whether two regions are globally similar, but they erase which feature dimensions carry the coherence or incoherence. DCA keeps this feature-dimension axis intact. This is why the output is a vector rather than one similarity score: the representational residual is distributed across dimensions, and averaging those dimensions destroys the signal.

First, DCA does not center the regional features. For RC, the mean activation of dimension $d$ across a region may be part of the coherence signal itself: if the same identity-related dimension activates in both eyes and mouth, its positive mean is evidence of cross-region consistency, not a bias to subtract. Reintroducing centering collapses CelebDF-v2 AUC from 0.9106 to 0.459.

Second, DCA does not apply per-dimension $L_2$ normalization. In a self-distilled feature space, some dimensions carry stronger identity load than others, and their magnitude is informative. Reintroducing per-dimension $L_2$ drops CelebDF-v2 AUC to 0.862, because strong identity-bearing dimensions are flattened toward weaker ones.

Third, DCA does not use the full cross-dimension Gram matrix. Off-diagonals ask whether dimension $d_i$ in one region interacts with $d_j$ in another, which is useful for bilinear aggregation but unrelated to RC. With $D^2 - D$ off-diagonal entries against $D$ diagonal entries, even weak off-diagonal noise can dominate the feature; reintroducing full coupling collapses CelebDF-v2 AUC to 0.478.

We call this principle \emph{constraint liberation}. We use the term descriptively: constraint liberation means removing normalization and coupling assumptions that are useful for cross-model comparison but mismatched to intra-sample measurement. The term does not mean that fewer operations are always better. It means that, in the RC regime, the usual constraints remove or obscure the quantities that carry coherence: regional mean activation, per-dimension magnitude, and same-dimension coactivation.

\begin{table}[t]
\centering
\caption{Constraint-liberation ablation on block-18 BN-normalized DINOv3 ViT-L/16 features. All rows use the same FF++ c23 training split, the same eyes-mouth-nose region set, and the same linear probe. Each ablation reintroduces one or more constraints that DCA removes.}
\label{tab:constraint_liberation}
\small
\begin{tabular}{lcccc}
\toprule
Variant & Centering & $L_2$ & Off-diag & CelebDF-v2 AUC \\
\midrule
\textbf{DCA} & -- & -- & -- & \textbf{0.9106} \\
\texttt{cosine\_dim} & -- & $\checkmark$ & -- & 0.862 \\
\texttt{cross\_covariance} & $\checkmark$ & -- & -- & 0.459 \\
\texttt{pnka\_dim} & $\checkmark$ & $\checkmark$ & $\checkmark$ & 0.478 \\
\bottomrule
\end{tabular}
\end{table}

The ablation pattern rules out a common interpretation of DCA as merely a simplified Gram or CKA variant. If DCA were only a weaker approximation of those operators, adding centering, normalization, or richer cross-dimension structure should not destroy the signal. Instead, each added constraint makes the measurement worse. The useful residual is specifically the raw same-dimension coactivation across semantic regions.

The remaining question is which frozen models support this measurement. DCA assumes that a feature dimension has a stable role across spatial positions: the same semantic factor should activate the same dimension whether it appears near the eyes, nose, or mouth. We hypothesize that DINOv3's self-distillation objective is a source of this property. By enforcing consistency across multiple crops of the same image, the model may be pressured to represent semantic content in a coordinate system that remains stable across spatial views.

This property is not guaranteed for every vision backbone. A model trained for supervised parsing, for example, may organize its features around label boundaries rather than identity-preserving geometry. We therefore treat backbone substitution as a diagnostic for the enabling condition. DINOv3 supports the RC measurement: face-interior regions show strong CKA-proxy agreement with DCA, with Pearson correlations above 0.99 for eyes, mouth, nose, and skin. FaRL does not support the same measurement: substituting FaRL ViT-B/16 for DINOv3 collapses CelebDF-v2 AUC to 0.582. This collapse is not a failure of the region pipeline; it indicates that the required coordinate-system property is absent or much weaker.

Hair makes the same boundary condition visible within DINOv3. Unlike eyes, mouth, and nose, hair is spatially variable and weakly tied to facial identity. The CKA-proxy correlation for hair collapses, and including hair in the region set substantially degrades cross-dataset performance. We therefore scope the headline DCA fingerprint to the eyes-mouth-nose triple. This is a principled scope choice: DCA should be applied to regions where the frozen model's coordinate system is stable enough for cross-region coherence to be measured.

Taken together, these results support the central measurement claim. RC is not obtained by adding a more expressive feature comparison. It appears when the comparison is restricted to the quantities that one frozen coordinate system makes meaningful: same-dimension activation, preserved magnitude, and cross-region coherence inside one input.

\section{Method}
\label{sec:method}

DCA measures whether semantic regions of a single face remain coherent inside a frozen foundation model. The pipeline has four stages: obtain a stable patch-token representation, assign tokens to semantic face regions, compute cross-region dimensional coactivation, and train a lightweight probe only for validation. The representation and region assignment modules are frozen throughout; no target-domain data is used at any stage.

\begin{figure}[!t]
\centering
\resizebox{0.95\linewidth}{!}{%
\begin{tikzpicture}[
  font=\rmfamily\footnotesize,
  node distance=4mm,
  inputbox/.style={rectangle, rounded corners=2pt, text centered, text width=2.0cm,
    minimum height=1.0cm, inner sep=2pt, draw=gray!60, fill=gray!10, line width=0.5pt},
  farlbox/.style={rectangle, rounded corners=2pt, text centered, text width=2.2cm,
    minimum height=1.0cm, inner sep=2pt, fill=farl-fill, draw=farl-draw, line width=0.5pt},
  dinobox/.style={rectangle, rounded corners=2pt, text centered, text width=2.2cm,
    minimum height=1.0cm, inner sep=2pt, fill=dino-fill, draw=dino-draw, line width=0.5pt},
  poolbox/.style={rectangle, rounded corners=2pt, text centered, text width=2.0cm,
    minimum height=1.0cm, inner sep=2pt, fill=rfg-fill, draw=rfg-draw, line width=0.5pt},
  dcabox/.style={rectangle, rounded corners=3pt, text centered, text width=2.4cm,
    minimum height=1.0cm, inner sep=3pt, fill=dca-fill, draw=dca-draw, line width=0.7pt},
  outputbox/.style={rectangle, rounded corners=3pt, text centered, text width=2.4cm,
    minimum height=1.0cm, inner sep=3pt, fill=real-fill, draw=real-draw, line width=0.9pt},
  mainarr/.style={->, >=Stealth, color=arrow-gray, line width=0.7pt},
]
\node[inputbox] (input) {%
  \textbf{Face Crop}\\[-1pt]
  {\scriptsize 448$\times$448}\\[-1pt]
  {\scriptsize RetinaFace}};
\node[dinobox, right=of input] (dino) {%
  \textbf{\textcolor{dino-draw}{DINOv3}}\\[-1pt]
  {\scriptsize ViT-L/16, frozen}\\[-1pt]
  {\scriptsize block 18 + BN}};
\node[farlbox, right=of dino] (farl) {%
  \textbf{\textcolor{farl-draw}{FaRL}}\\[-1pt]
  {\scriptsize lapa/448}\\[-1pt]
  {\scriptsize per-patch labels}};
\node[poolbox, right=of farl] (pool) {%
  \textbf{\textcolor{rfg-draw}{EMN pool}}\\[-1pt]
  {\scriptsize eyes, mouth,}\\[-1pt]
  {\scriptsize nose; $K{=}20$}};
\node[dcabox, right=of pool] (dca) {%
  \textbf{\textcolor{dca-draw}{DCA}}\\[-1pt]
  {\scriptsize $\frac{1}{K}\,\mathrm{diag}(F_{r_1}^{\!\top} F_{r_2})$}\\[-1pt]
  {\scriptsize 3 pairs $\cdot$ 1024}};
\node[poolbox, right=of dca] (probe) {%
  \textbf{\textcolor{rfg-draw}{LR probe}}\\[-1pt]
  {\scriptsize $\phi \in \mathbb{R}^{3072}$}\\[-1pt]
  {\scriptsize trained on FF++}};
\node[outputbox, right=of probe] (output) {%
  \textbf{\textcolor{real-draw}{$\hat{y}$}}\\[-1pt]
  {\scriptsize CDF\,=\,0.911}\\[-1pt]
  {\scriptsize DFD\,=\,0.929}};
\draw[mainarr] (input) -- (dino);
\draw[mainarr] (dino) -- (farl);
\draw[mainarr] (farl) -- (pool);
\draw[mainarr] (pool) -- (dca);
\draw[mainarr] (dca) -- (probe);
\draw[mainarr] (probe) -- (output);
\end{tikzpicture}%
}
\caption{DCA pipeline. A face frame is converted into frozen DINOv3 patch tokens, assigned to semantic regions using a frozen FaRL parser, and transformed into a 3{,}072-dimensional eyes-mouth-nose DCA fingerprint. The fingerprint measures same-dimension coactivation across region pairs and is evaluated with a linear probe trained only on FF++ c23. No backbone fine-tuning, target-domain data, within-sample centering, $L_2$ normalization, or Gram off-diagonals are used.}
\label{fig:fig1}
\end{figure}
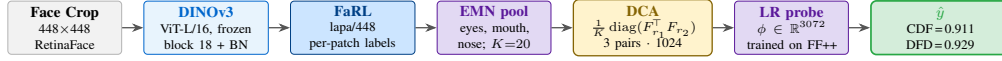

\paragraph{Frozen patch-token representation.}
Each video frame is first detected with RetinaFace~\citep{deng2020retinaface}, padded by 0.22, and resized to $448 \times 448$. A frozen DINOv3 ViT-L/16 backbone~\citep{simeoni2025dinov3} then maps the face crop to a $28 \times 28$ grid of patch tokens, $\mathbf{X} \in \mathbb{R}^{784 \times 1024}$. We use intermediate block-18 activations rather than the final normalized patch tokens because the intermediate representation preserves richer per-dimension structure. The block index is 1-indexed in the experiment manifest and corresponds to PyTorch block 17.

Intermediate tokens are extracted with \texttt{norm=False}. We then apply per-dimension batch normalization using statistics computed only from the FF++ training split:
\[
\hat{x}_d = \frac{x_d - \mu_d}{\sigma_d}.
\]
This normalization is not the same as the centering ablated in Section~\ref{sec:rc_constraint_liberation}. The batch-normalization statistics are fixed from the source training split and are applied once to stabilize the intermediate-layer feature scale. DCA itself performs no within-sample centering, no $L_2$ normalization, and no cross-dimension coupling.

\paragraph{Semantic region assignment.}
To compare regions, each patch token must be assigned an anatomical label. We use a frozen FaRL ViT-B/16 face parser~\citep{zheng2022farl} trained on LaPa labels to produce a $448 \times 448$ segmentation map. FaRL is used only as a region-assignment module, not as the representation backbone for DCA. Each DINOv3 patch is assigned the label at its center pixel. The original parsing labels are merged into five anatomical regions: eyes, mouth, nose, skin, and hair.

The headline fingerprint uses only the eyes-mouth-nose (EMN) region triple. This choice follows the scope argument in Section~\ref{sec:rc_constraint_liberation}: these face-interior regions are the most stable carriers of the DINOv3 coordinate-system property. Hair is excluded because its coordinate-system agreement collapses, and skin is excluded from the headline fingerprint because adding skin pairs empirically dilutes the cross-dataset signal.

\paragraph{Patch sampling.}
For each retained region, we sample $K=20$ patch tokens. If a region contains fewer than $K$ tokens, sampling is performed with replacement. This produces a fixed-size matrix $F_r \in \mathbb{R}^{K \times D}$ for every region $r$, with $D=1024$. The shared sample index $k$ in the DCA formula is an index over sampled patches, not a claim of spatial correspondence between regions. DCA does not require the $k$-th eye patch to correspond geometrically to the $k$-th mouth patch; it measures whether the same feature dimensions coactivate across the sampled regional token sets.

\paragraph{DCA fingerprint.}
For each unordered pair in the EMN triple,
\[
(\text{eyes}, \text{mouth}), \quad
(\text{eyes}, \text{nose}), \quad
(\text{mouth}, \text{nose}),
\]
we compute the DCA vector (Eq.~\ref{eq:dca}):
\[
\mathrm{DCA}(F_{r_1},F_{r_2})
=
\frac{1}{K}\,\mathrm{diag}(F_{r_1}^{\top}F_{r_2})
\in \mathbb{R}^{1024}.
\]
The three vectors are concatenated to form the sample-level fingerprint:
\[
\phi =
\bigl[\,
\mathrm{DCA}(F_{\text{eyes}},F_{\text{mouth}})
\;\|\;
\mathrm{DCA}(F_{\text{eyes}},F_{\text{nose}})
\;\|\;
\mathrm{DCA}(F_{\text{mouth}},F_{\text{nose}})
\,\bigr]
\in \mathbb{R}^{3072}.
\]
This fingerprint is the feature-dimension-preserving RC measurement used throughout the main experiments. It is parameter-free and is computed entirely from frozen features.

\paragraph{Validation probe.}
The classifier is intentionally simple: a logistic-regression probe trained on FF++ c23 only. The probe uses class-balanced weights, $C=0.1$, the \texttt{lbfgs} solver, \texttt{max\_iter=2000}, and \texttt{random\_state=42}. A \texttt{StandardScaler} is fit on the FF++ training fingerprints only and then applied unchanged to all evaluation sets. The probe is used to validate whether the DCA fingerprint carries cross-dataset signal; it is not part of the RC measurement itself.

\paragraph{Video-level scoring.}
For each video, we sample 15 evenly spaced frames and compute one DCA fingerprint per frame. The trained probe produces a frame-level score, and video-level prediction is obtained by mean-pooling the 15 frame scores. We report video-level ROC-AUC with 95\% bootstrap confidence intervals using 1{,}000 resamples and seed 42.

\paragraph{Implementation summary.}
The full method can be summarized as:
\[
\text{frame}
\rightarrow
\text{face crop}
\rightarrow
\text{DINOv3 block-18 tokens}
\rightarrow
\text{FaRL region assignment}
\rightarrow
\text{EMN patch sampling}
\rightarrow
\text{DCA pair vectors}
\rightarrow
\text{3{,}072-d fingerprint}
\rightarrow
\text{linear probe}.
\]
All representation modules are frozen. The only learned component is the final logistic-regression probe used for validation.

\paragraph{Reproducibility.}
All reported DCA features are extracted from frozen DINOv3 ViT-L/16 intermediate block-18 tokens with \texttt{norm=False}. Batch-normalization statistics are computed only from the FF++ training split and will be released with the code. The FaRL parser is used only for region assignment. We will release the feature-extraction scripts, BN statistics, split files, and logistic-regression evaluation code. The final version will include the exact DINOv3 checkpoint identifier and pinned package versions.

\section{Experiments and Results}
\label{sec:results}

We use cross-dataset deepfake detection as the validation task for Representational Consistency. The task is appropriate because it directly tests whether a measurement learned from one distribution transfers to unseen manipulation pipelines and identities. A detector that only memorizes source-domain artifacts should fail under this setting; a measurement of intra-sample coherence should transfer when the same representational failure appears across datasets.

All probes are trained on FaceForensics++ (FF++) c23~\citep{rossler2019ff} and evaluated zero-shot on CelebDF-v2~\citep{li2020celebdf} and DeepFakeDetection (DFD)~\citep{google2019dfd}. No target-domain labels, features, or calibration data are used. Unless otherwise stated, all results use block-18 BN-normalized DINOv3 ViT-L/16 patch tokens, the eyes-mouth-nose (EMN) region triple, and the direct logistic-regression probe described in Section~\ref{sec:method}.

\paragraph{Datasets and protocol.}
FF++ c23 provides the source training distribution, with real videos and four manipulation types. CelebDF-v2 and DFD are used only for zero-shot evaluation. For each video, we evaluate 15 evenly spaced frames and mean-pool frame-level scores into a video-level prediction. We report video-level ROC-AUC with 95\% bootstrap confidence intervals computed from 1{,}000 resamples.

\paragraph{Cross-dataset validation of DCA.}
Table~\ref{tab:main_results} compares DCA against regional pooled baselines, scalar correspondence reductions, global DINOv3 features, and backbone substitutions. These comparisons are intended to isolate the effect of the RC measurement under a shared backbone and protocol, not to claim a reproduced comparison against the full deepfake-detection literature. The DCA EMN fingerprint achieves 0.9106 AUC on CelebDF-v2 and 0.9289 AUC on DFD under FF++ c23 $\rightarrow$ target transfer. On CelebDF-v2, DCA outperforms all single-region pooled baselines and the strongest pooled multi-region baseline. On DFD, DCA remains competitive but does not uniformly dominate: the mouth+skin pooled baseline is higher by approximately 1.0 pp. We therefore frame the result carefully. DCA is not a claim of universal detector dominance; it is evidence that preserving cross-region per-dimension coherence exposes a transferable signal that single-region and scalar measurements miss.

\begin{table}[t]
\centering
\caption{Main cross-dataset result. All methods are trained on FF++ c23 and evaluated zero-shot. DCA uses the eyes-mouth-nose region triple and produces a 3{,}072-dimensional fingerprint. The strongest result on CelebDF-v2 validates DCA as an RC measurement; DFD shows competitive but not uniform dominance.}
\label{tab:main_results}
\small
\begin{tabular}{lccc}
\toprule
Method & FF++ Test & CelebDF-v2 & DFD \\
\midrule
\textbf{DCA EMN (3{,}072 dims)} & \textbf{0.989} & \textbf{0.9106} & 0.9289 \\
A1 single region: nose & 0.958 & 0.8886 & 0.9299 \\
A1 single region: mouth & 0.963 & 0.8580 & 0.9299 \\
A1 single region: eyes & 0.947 & 0.8310 & 0.8870 \\
A1 multi-region: mouth $+$ skin & 0.969 & 0.8991 & \textbf{0.9391} \\
DCA scalar reductions & --- & 0.467--0.615 & 0.547--0.661 \\
Cross-region Gram Frobenius & 0.560 & 0.427 & 0.629 \\
DINOv2 ViT-L/14 DCA EMN & 0.983 & 0.785 & 0.849 \\
DINOv3 CLS global & 0.884 & 0.641 & 0.805 \\
FaRL ViT-B/16 backbone $+$ EMN pipeline & 0.930 & 0.582 & 0.757 \\
\bottomrule
\end{tabular}
\end{table}

The CelebDF-v2 result is the main cross-dataset validation. CelebDF-v2 is a harder target because its manipulations are less tied to the FF++ artifact distribution. DCA improves over the strongest single-region baseline (nose, 0.8886) by $+$2.2 pp and over the strongest pooled baseline (mouth$+$skin, 0.8991) by $+$1.15 pp. The latter gap is modest, so we do not overstate it as decisive superiority. The stronger claim is that DCA reaches the best CelebDF-v2 transfer while using a principled cross-region measurement rather than a pooled regional descriptor.

On DFD, pooled region baselines (mouth, nose, mouth$+$skin) are already strong; DCA remains within a narrow range but does not exceed the best, consistent with DFD's artifacts being well captured by local regional pooling.

\paragraph{Constraint liberation is load-bearing.}
The operator-level ablation in Table~\ref{tab:constraint_liberation} shows that DCA's design choices are load-bearing. Reintroducing centering, $L_2$ normalization, or cross-dimension coupling damages the signal, with centering and full Gram coupling collapsing CelebDF-v2 AUC to near chance.

\paragraph{The signal lives on the feature-dimension axis.}
The scalar baselines test whether RC can be reduced to one similarity score per region pair. It cannot. Cosine similarity over region means, patch-level CKA, mean DCA, and nearest-neighbor cosine all fall into the 0.47--0.66 AUC range. A cross-region Gram Frobenius baseline also fails, reaching 0.427 AUC on CelebDF-v2. These failures are informative: the regions are not unrelated, but the discriminative information is not captured by their average similarity. It is distributed across feature dimensions.

This supports the need to preserve the feature-dimension axis. Scalar summaries collapse heterogeneous dimensions into one number and erase which dimensions carry coherence or incoherence. DCA preserves the feature-dimension axis, making the frozen coordinate system readable. The jump from scalar reductions to the 3{,}072-dimensional DCA fingerprint is therefore not merely a dimensionality advantage; it is a change in what is being measured.

\paragraph{DCA depends on the backbone's coordinate system.}
Replacing DINOv3 with FaRL ViT-B/16 while keeping the same region pipeline collapses CelebDF-v2 AUC to 0.582. This is consistent with FaRL's supervised parsing objective: it optimizes for categorical region boundaries rather than a stable identity-preserving coordinate system. DINOv2~\citep{oquab2023dinov2} gives an intermediate 0.785 AUC: self-distillation helps, but coordinate-system precision improves with pretraining scale. The DINOv3 CLS global baseline (0.641) shows that the signal is not available in the global image representation; it appears only when semantic regions are compared through the per-dimension axis.

\paragraph{Block selection and feature depth.}
DCA is strongest at an intermediate DINOv3 block rather than the final layer. Table~\ref{tab:block_sweep} reports the block sweep. Block 18 reaches the highest CelebDF-v2 and DFD AUC, while the final normalized patch tokens are lower. This suggests that intermediate features preserve richer per-dimension structure for RC measurement, whereas final features are more compressed by the end of the model.

\begin{table}[t]
\centering
\caption{DCA EMN performance across DINOv3 ViT-L/16 blocks. Intermediate block-18 features provide the strongest cross-dataset transfer. Final \texttt{x\_norm\_patchtokens} underperform on CelebDF-v2, which suggests that late normalization and abstraction reduce the per-dimension structure needed for RC measurement.}
\label{tab:block_sweep}
\small
\begin{tabular}{lccc}
\toprule
Model / Block & FF++ Test & CelebDF-v2 & DFD \\
\midrule
DINOv3 block 16 & 0.986 & 0.887 & 0.890 \\
\textbf{DINOv3 block 18} & \textbf{0.989} & \textbf{0.9106} & \textbf{0.9289} \\
DINOv3 block 20 & 0.973 & 0.843 & 0.897 \\
DINOv3 block 22 & 0.966 & 0.854 & 0.880 \\
DINOv3 block 24 & 0.950 & 0.807 & 0.878 \\
DINOv3 final patch tokens & 0.961 & 0.8967 & 0.917 \\
CLIP ViT-L/14 final & 0.881 & 0.668 & 0.790 \\
\bottomrule
\end{tabular}
\end{table}

Because this block was selected from a target-domain sweep, we treat the block result as an empirical diagnostic rather than as a source of additional claimed improvement; the stronger conclusion is that RC measurement is sensitive to where the frozen coordinate system is read.

\paragraph{Region-set ablation.}
Adding more semantic regions does not improve DCA. The EMN triple is strongest on CelebDF-v2; adding skin reduces AUC to 0.8902, and including all five regions (with hair) collapses CelebDF-v2 AUC to 0.751. RC measurement should use regions where the coordinate-system property is stable; unstable regions inject incoherent profiles that dilute the measurement (full table in Appendix~\ref{app:region_set_ablation}).

\section{Discussion and Future Work}
\label{sec:discussion}

Frozen foundation models with a stable coordinate system are more than feature extractors: their internal representation is a reference for whether a single input is coherent across semantic regions. The main claim is not that DCA is a universally stronger detector. On DFD, strong pooled region baselines remain competitive. The claim is that cross-region representational coherence is a real and measurable signal that scalar summaries and region-pooled baselines miss on the harder CelebDF-v2 transfer.

\paragraph{What DCA reveals.}
DCA reveals a residual that is relational, feature-dimensional, and backbone-dependent. It is relational because it compares whether regions remain mutually coherent rather than whether one region is locally abnormal. It is feature-dimensional because scalar summaries erase which dimensions carry coherence or incoherence. It is backbone-dependent because the signal appears clearly in DINOv3 but collapses under FaRL, which indicates that the measurement requires a stable per-dimension coordinate system.

\paragraph{Boundary conditions.}
The region and manipulation analyses identify where this measurement is weaker. Hair is spatially variable and weakly tied to identity, so it injects incoherent profiles rather than useful cross-region signal. NeuralTextures is a different boundary case: globally diffuse texture transfer may not create sharp single-frame cross-region residuals. These failures clarify the scope of DCA rather than contradicting the RC framing.

Several limitations remain. The current validation is face-specific; extending RC to general objects requires reliable semantic part discovery. The benchmark suite is limited to established deepfake datasets; broader validation should include DFDC, WildDeepFake, FFIW, and modern diffusion-based manipulations. Results use a single probe seed and a block selected by an OOD sweep; future versions should select the block on a held-out source validation split, report multi-seed variance, and include paired statistical tests. We do not evaluate demographic robustness, which deepfake detectors are known to fail at unevenly. This paper focuses on measurement; per-region DCA scores may support future applications that respond differently depending on where coherence breaks, but those applications require separate validation.

\section{Conclusion}
\label{sec:conclusion}

We introduced Representational Consistency, the problem of measuring whether a frozen foundation model represents a single input as internally coherent across its semantic subregions, and instantiated it with Dimensional Coactivation (DCA), a per-dimension operator over patch-token features. The coherence signal becomes measurable only when the feature-dimension axis is preserved: scalar summaries erase the residual, and standard similarity constraints (centering, $L_2$ normalization, and off-diagonal Gram coupling) each damage the signal.

On frozen DINOv3, the eyes-mouth-nose DCA fingerprint transfers zero-shot from FF++ c23 to CelebDF-v2 (0.9106) and DFD (0.9289), and the FaRL substitution shows the measurement depends on the backbone's coordinate system rather than the segmentation pipeline. Deepfake detection is the first validation. More broadly, frozen foundation models can act as reference systems that detect when generated, corrupted, or modified inputs break physical-world structure.

\paragraph{Broader impact.}
Deepfake detection carries dual-use risk: public detection methods can inform adversarial generators or give users false confidence if deployed without validation. We mitigate by framing the contribution as a measurement principle, not a deployable forensic model. The method should not be used for real-world authenticity judgments without broader validation, especially against modern diffusion-based manipulations.

\bibliographystyle{abbrvnat}
\bibliography{refs}

\appendix

\section{Additional Constraint Ablations}
\label{app:constraint_ablations}

Section~\ref{sec:rc_constraint_liberation} showed that DCA depends on removing three constraints that are standard in representation similarity: within-sample centering, per-dimension $L_2$ normalization, and off-diagonal cross-dimension coupling. This appendix provides additional intuition for why these operations damage the RC signal.

\paragraph{Centering removes the regional mean signal.}
Classical similarity measures often center feature matrices before computing a similarity score. This is appropriate when the feature mean is treated as an arbitrary offset. In RC measurement, however, the regional mean activation of a feature dimension may carry coherence signal. If dimension $d$ is jointly activated in both eyes and mouth, its positive mean across sampled patches is evidence that the two regions are coordinated inside the frozen model's coordinate system.

When we reintroduce centering, CelebDF-v2 AUC collapses from 0.9106 to 0.459. This indicates that the removed mean is not nuisance variation. It is part of the signal DCA is designed to preserve.

\paragraph{$L_2$ normalization weakens magnitude information.}
Per-dimension $L_2$ normalization makes each dimension contribute with equal length. This is useful when scale is considered arbitrary, but DCA assumes that magnitude can be semantically loaded. In a self-distilled feature space, some dimensions may carry stronger identity or region-coherence signal than others. Normalizing these dimensions weakens the difference between high-signal and low-signal axes.

Reintroducing per-dimension $L_2$ normalization reduces CelebDF-v2 AUC from 0.9106 to 0.862. The drop is smaller than the centering collapse, but it shows that magnitude is not merely scale noise in this measurement regime.

\paragraph{Off-diagonal coupling overwhelms the diagonal residual.}
Full Gram-style operators include all interactions between dimension $d_i$ in one region and dimension $d_j$ in another region. DCA keeps only the diagonal, where the same feature dimension coactivates across both regions. This diagonal has a direct coordinate-system interpretation: it asks whether dimension $d$ remains coherent across semantic regions.

The off-diagonal terms are much more numerous. For $D=1024$, the diagonal has 1{,}024 entries, while the off-diagonal part has more than one million entries. Even if each off-diagonal entry is only weakly noisy, the total volume can overwhelm the same-dimension residual. Reintroducing full cross-dimension coupling collapses CelebDF-v2 AUC to 0.478. The corresponding ablation table is reported in the main text (Table~\ref{tab:constraint_liberation}).

These ablations support the central design choice of DCA. The useful residual is not created by adding a richer similarity operator. It is exposed by preserving raw same-dimension coactivation across semantic regions.

\section{Block Sweep}
\label{app:block_sweep}

DCA is sensitive to where the frozen coordinate system is read. We therefore evaluate the same eyes-mouth-nose DCA pipeline across several DINOv3 ViT-L/16 blocks. Intermediate block-18 features provide the strongest cross-dataset transfer.

\begin{table}[h]
\centering
\caption{DCA EMN performance across DINOv3 ViT-L/16 blocks. Intermediate features provide the strongest transfer. Final \texttt{x\_norm\_patchtokens} underperform on CelebDF-v2, which suggests that late normalization and compression reduce the per-dimension structure needed for RC measurement.}
\label{tab:appendix_block_sweep}
\small
\begin{tabular}{lccc}
\toprule
Model / Block & FF++ Test & CelebDF-v2 & DFD \\
\midrule
DINOv3 block 16 & 0.986 & 0.887 & 0.890 \\
\textbf{DINOv3 block 18} & \textbf{0.989} & \textbf{0.9106} & \textbf{0.9289} \\
DINOv3 block 20 & 0.973 & 0.843 & 0.897 \\
DINOv3 block 22 & 0.966 & 0.854 & 0.880 \\
DINOv3 block 24 & 0.950 & 0.807 & 0.878 \\
DINOv3 final patch tokens & 0.961 & 0.8967 & 0.917 \\
CLIP ViT-L/14 final & 0.881 & 0.668 & 0.790 \\
\bottomrule
\end{tabular}
\end{table}

Block 18 is the strongest point in this sweep. We treat this as empirical model selection rather than a universal claim about layer depth. The result suggests that RC measurement benefits from features that are semantically organized but not yet fully compressed by the final representation.

A limitation is that the current block choice is selected from an OOD sweep. A stricter future protocol should select the block using only a held-out source-domain validation split and then report target-domain results once.

\section{Region-Set Ablation}
\label{app:region_set_ablation}

DCA does not improve monotonically as more regions are added. The eyes-mouth-nose region triple is the strongest cross-dataset configuration. Adding skin increases dimensionality but reduces CelebDF-v2 performance, and including hair sharply degrades transfer.

\begin{table}[h]
\centering
\caption{Region-set ablation for DCA. All variants use block-18 BN-normalized DINOv3 features and the same direct linear-probe protocol. EMN is the strongest cross-dataset region set. Adding skin dilutes the signal; adding hair strongly degrades performance.}
\label{tab:appendix_region_ablation}
\small
\begin{tabular}{lcccc}
\toprule
Region set & Pairs $\times$ 1024 & FF++ Test & CelebDF-v2 & DFD \\
\midrule
\textbf{Eyes, mouth, nose} & \textbf{3{,}072} & \textbf{0.989} & \textbf{0.9106} & \textbf{0.9289} \\
Eyes, mouth, nose, skin & 6{,}144 & 0.969 & 0.8902 & 0.9244 \\
Eyes, mouth & 1{,}024 & 0.964 & 0.881 & 0.917 \\
All five regions including hair & 10{,}240 & 0.927 & 0.751 & 0.802 \\
\bottomrule
\end{tabular}
\end{table}

This ablation supports the interpretation that region choice is part of the measurement definition. DCA should be applied to regions where the frozen model maintains a stable coordinate system. Eyes, mouth, and nose are stable face-interior anchors. Hair is spatially variable and weakly tied to facial identity, so it injects incoherent profiles rather than useful cross-region signal.

\section{Scalar Reductions}
\label{app:scalar_reductions}

Scalar summaries are structurally insufficient on this task because they collapse the feature-dimension axis where the discriminative residual lives. We evaluate several scalar alternatives using the same region pipeline and the same source-domain training protocol. All scalar reductions perform far below the per-dimension DCA fingerprint.

\begin{table}[h]
\centering
\caption{Scalar correspondence reductions on the same DINOv3 region pipeline. Each method collapses region-pair information to scalar summaries before classification. The collapse in performance shows that the discriminative residual is distributed across feature dimensions and is erased by scalar averaging.}
\label{tab:appendix_scalar_reductions}
\small
\begin{tabular}{lccc}
\toprule
Reduction & FF++ Test & CelebDF-v2 & DFD \\
\midrule
Cosine on region means & 0.573 & 0.615 & 0.661 \\
CKA on patches & 0.488 & 0.573 & 0.591 \\
DCA mean over dimensions & 0.467 & 0.547 & 0.612 \\
Cross-region nearest-neighbor cosine & 0.572 & 0.538 & 0.577 \\
\textbf{DCA per-dimension EMN} & \textbf{0.989} & \textbf{0.9106} & \textbf{0.9289} \\
\bottomrule
\end{tabular}
\end{table}

These results show that the regions are not simply more or less similar as wholes. The signal is distributed unevenly across feature dimensions. DCA works because it preserves which dimensions coactivate across regions instead of reducing the comparison to one similarity score.

\end{document}